\documentclass{article}

\usepackage[preprint]{neurips_2025}

\usepackage{microtype}
\usepackage{graphicx}
\usepackage{subfigure}
\usepackage{booktabs} 
\usepackage{multirow}
\usepackage{hyperref}
\usepackage{algorithm,algorithmic}
\usepackage{colortbl} 
\usepackage{xcolor} 
\usepackage{amsmath}
\usepackage{amssymb}
\usepackage{mathtools}
\usepackage{amsthm}
\usepackage[utf8]{inputenc} 
\usepackage[T1]{fontenc}    
\usepackage{url}            
\usepackage{amsfonts}       
\usepackage{nicefrac}       
\usepackage{natbib}

\usepackage[capitalize,noabbrev]{cleveref}

\theoremstyle{plain}

\theoremstyle{definition}

\theoremstyle{remark}

\usepackage{wrapfig}

\usepackage[textsize=tiny]{todonotes}

\title{Partial GFlowNet: Accelerating Convergence in Large State Spaces via Strategic Partitioning}

\author{
  Xuan Yu \\
  \texttt{yx2024@mail.ustc.edu.cn} \\
  University of Science and Technology of China (USTC), Hefei, China \\
  \And
  Ruiqi Wu \\
  Inner Mongolia Minzu University (IMUN), Tongliao, China \\
  \And
  Xu Wang\thanks{Corresponding authors: \texttt{wx309@ustc.edu.cn}} \\
  University of Science and Technology of China (USTC), Hefei, China \\
  Suzhou Institute for Advanced Research, USTC, Suzhou, China \\
  \And
  Rui Zhu \\
  University of Science and Technology of China (USTC), Hefei, China \\
  \And
  Yudong Zhang \\
  Suzhou Institute for Advanced Research, USTC, Suzhou, China \\
  \And
  Yang Wang\thanks{Corresponding authors: \texttt{angyan@ustc.edu.cn}} \\
  University of Science and Technology of China (USTC), Hefei, China \\
  State Key Laboratory of Precision and Intelligent Chemistry, USTC, Hefei, China \\
}

\begin{document}

\maketitle

\newcommand{\modelname}{{Partial GFlowNet }}
\newcommand{\modelnames}{{Partial GFlowNets }}
\newcommand{\modelnameNoBlank}{{Partial GFlowNet}}
\newcommand{\gfn}{{GFlowNet}}
\newcommand{\gfns}{{GFlowNets}}
\newcommand{\lsgfn}{{Local Search GFlowNet}}

\begin{abstract}
Generative Flow Networks (GFlowNets) have shown promising potential to generate high-scoring candidates with probability proportional to their rewards. As existing GFlowNets freely explore in state space, they encounter significant convergence challenges when scaling to large state spaces. Addressing this issue, this paper proposes to restrict the exploration of actor. A planner is introduced to partition the entire state space into overlapping partial state spaces. Given their limited size, these partial state spaces allow the actor to efficiently identify subregions with higher rewards. A heuristic strategy is introduced to switch partial regions thus preventing the actor from wasting time exploring fully explored or low-reward partial regions. By iteratively exploring these partial state spaces, the actor learns to converge towards the high-reward subregions within the entire state space. Experiments on several widely used datasets demonstrate that \modelname converges faster than existing works on large state spaces. Furthermore, \modelname not only generates candidates with higher rewards but also significantly improves their diversity.
\end{abstract}

\section{Introduction} \label{sec:intro}

Generative Flow Networks (GFlowNets; GFNs)~\cite{bengio2021flow} have demonstrated remarkable potential in generating diverse and high-reward candidates. These networks learn to sample candidates with probability proportional to their rewards. 
Typical candidates generated by GFNs are the terminal of state trajectories, where the transitions of states are controlled by the action sequences of actor. Specifically, the actor evaluates all possible actions from the current state and predicts the probability of each possible action. It chooses an action based on a learned forward policy. Such procedure ensures that actions leading to high-reward outcomes are more likely to be chosen. 
\begin{wrapfigure}{r}{0.35\textwidth}
    \label{fig:flow-graph}
    \includegraphics[width=0.35\textwidth]{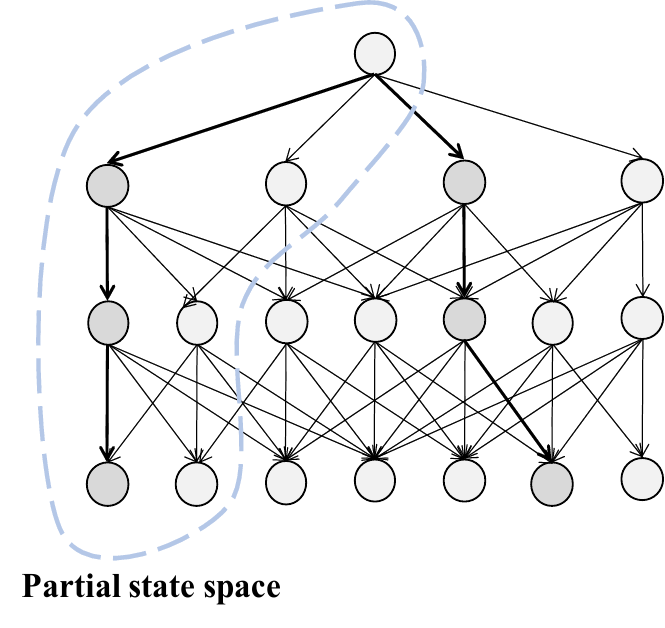}
    \caption{High-reward candidates can be rather sparse. By selecting partial space, the high-reward candidates become denser.}
\end{wrapfigure}
To predict the probabilities of possible actions, GFNs estimate flows in a flow graph, where intermediate nodes represent states and terminal nodes correspond to final candidates. This flow graph captures all possible state transitions during the generation process. The flows within the graph are guided by the rewards of terminal states, using strategies such as flow matching and other optimization techniques to ensure consistency and alignment with the reward function~\cite{bengio2021flow,malkin2022trajectory}. Edges, i.e., actions, with higher flows are more likely to be chosen. The key target of training GFlowNets is to learn a flow graph with proper flows within edges, thus enabling the generation of diverse and high-reward candidates.

As the flow graph is built upon state transitions, the size of flow graph is correlated to the number of possible actions and the length of trajectories. 
Fig.~\ref{fig:flow-graph} illuatrates that the size of the flow graph has an exponential relationship with the trajectory length and a power-law relationship with the number of actions. 
When the size of the flow graph in GFlowNets becomes very large, it is challenging to assign proper flows of edges and to search for high-reward subspace. 
In practice, in the task of generating molecules proposed by \cite{bengio2021flow}, the maximum trajectory length is set to $8$, with more than $105$ actions and number of possible trajectories exceeding $10^{16}$. Under these constraints, GFlowNets are capable of identifying over $10000$ distinct modes with rewards greater than $7.5$. If the maximum trajectory length is extended to $12$, GFlowNets should theoretically be able to discover more high-scoring candidates and modes, since the original state space $\mathcal{S}$ is a subset of the new state space $\mathcal{S}'$, i.e., $\mathcal{S} \subset \mathcal{S}'$. However, existing GFlowNets fail to discover as many high-scoring modes as expected. We test various methods across different seeds and find that they struggle to converge towards high-scoring regions, identifying fewer than $10$ modes with $R>7.5$, even after sampling over $10^6$ candidates. 

This is an exploration challenge in extremely large spaces. As the high-reward areas are rather sparse, GFlowNets become stuck in low-reward areas and struggle to transition to high-reward areas, essentially losing the ability to effectively explore high-reward areas and generate high-scoring candidates. 
What is favorable for GFlowNets is to thoroughly explore a partial region to identify high-scoring candidates, and then gradually transition from this region to others, step by step. Hence, we propose \modelname to guide GFlowNets in generating diverse, high-scoring candidates within extremely large state spaces.

In our proposed \modelname, we avoid directly exploring the entire state space $\mathcal{S}$ by partitioning it into distinct overlapping regions $\mathcal{R}$. We design a planner $P$ to guide the selection of $\mathcal{R}$. When sampling candidates through the actor of \modelnameNoBlank, these candidates are also utilized by the planner to update the policy for choosing the next region, $\mathcal{R}'$. $P$ evolves throughout the sampling process of the actor, and the actor is directed to focus on specific regions according to the guidance of $P$. The collaboration between the actor and planner $P$ allows \modelname to shift toward high-scoring regions within the state space, resulting in faster convergence compared to vanilla \gfns.

The main contributions of this work are:
\begin{itemize}
    \item Our research endeavors to adapt GFlowNets for large state space scenarios. The proposed \modelname adopts a heuristic algorithm to iteratively select subregions thus facilitating the exploration of actor of GFlownets.
    \item To enhance the quality of generated candidates, we introduce a partial local search algorithm. By modifying the local search algorithm process to suit large space, we achieve the final training algorithm of \modelnameNoBlank, termed Partial Local Search(PLS).
    \item The proposed method demonstrates impressive performance on benchmarks with large state space, and outperforms existing GFlowNet methods even on some standard benchmarks.
\end{itemize}

\section{Methodology} \label{sec:method}

\subsection{Problem formulation}
We describe the definition of the problem, where the goal is to train a policy to generate candidate objects $ x \in \mathcal{X} $ with probabilities proportional to a reward function $ R(x): x \to \mathbb{R}^+ $. The process begins with an initial state $ s_0 $, from which a series of actions are taken to transform the state into $ x $. This process can be represented as a trajectory of state transformations, denoted as $ \tau = (s_0, s_1, s_2, \cdots, s_n = x) $. The set of all states is denoted as $ \mathcal{S} $, and the set of actions is given by $ \mathcal{A} = \{(s \to s') \mid s, s' \in \mathcal{S}\} $. 
Furthermore, we define $\mathcal{A}^*=(a_1^*,a_2^*,\cdots,a_n^*)$ as the part of actions irrelevant to the states and $\mathcal{A}'$ as the relevant part. For each $a \in \mathcal{A}$, $a $ can be decomposed into $a = (a^*,a')$ where $a^* \in \mathcal{A}^*$ and $a' \in \mathcal{A}'$.

It is important to note that we assume a one-to-one correspondence between actions and the resulting states, meaning that for each pair of states $ s $ and $ s' $, there exists exactly one action $ (s \to s') $ that transitions $ s $ to $ s' $. We define $ s $ as the parent of $ s' $ and $ s' $ as the child of $ s $ when $ (s \to s') \in \mathcal{A} $. Specifically, we define $ \mathcal{A}(s) $ as the set of actions that transition $ s $ to all its children, and for any terminal state $ x $, we have $ \mathcal{A}(x) = \varnothing $.

\begin{wrapfigure}{r}{0.4\textwidth}
    \label{fig:pipeline}
        \includegraphics[width=0.4\textwidth]{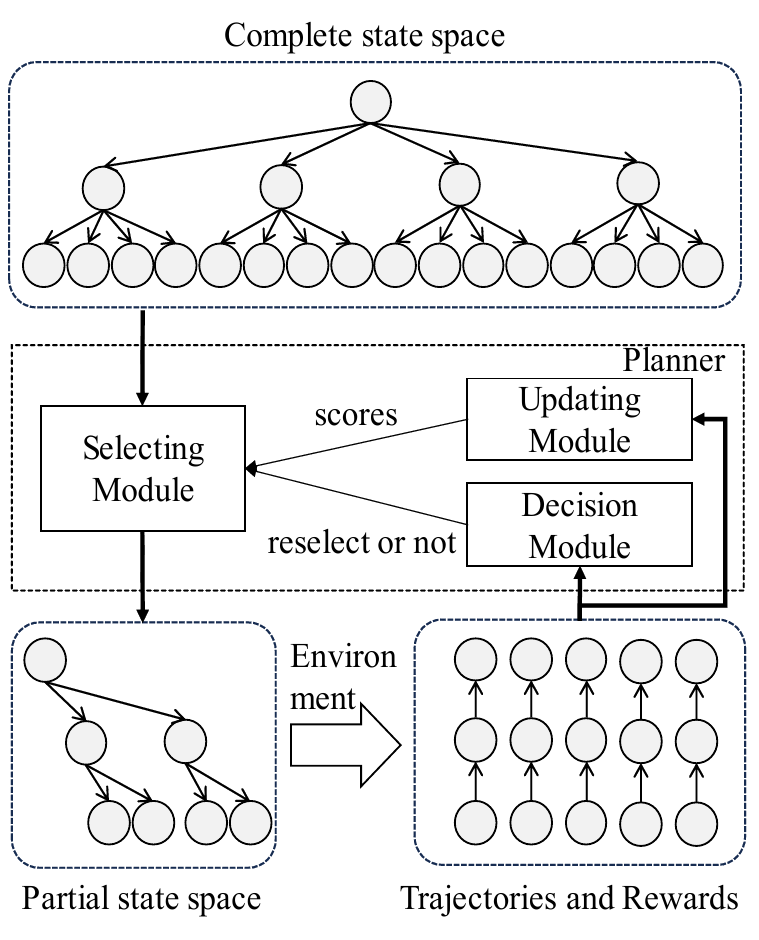}
        \caption{Overview of the proposed Partial GFlowNet framework.}
\end{wrapfigure}

\subsection{Partial Search}
\par Although GFlowNet is proposed to generate samples proportional to their corresponding rewards, it is not always effective. Such challenges arise especially in situations where the state space is large, where it is difficult for GFlowNet to identify high-scoring terminal states, i.e., candidates. 
To address this issue, we propose \modelnameNoBlank, which constrains the behavior of GFlowNet by dividing the whole state space into distinct partial regions, which may overlap or not. In \modelnameNoBlank, the actor is restricted to visit a designated partial region, unless it is instructed to switch to another region by the planner. The planner determines when the transition occurs and specifies the next partial region, as shown in Fig.\ref{fig:pipeline}.


To create distinct partial regions, we impose constraints on $\mathcal{A}^*$, as they are irrelevant to states. Specifically, we sample actions from $\mathcal{A}^*$ with the strategy, where an action $a \in \mathcal{A}^*$ is considered valid with probability $p$ and invalid with probability $1-p$.
This sampling process yields a valid action set $\mathcal{A}^*_v$, which remains consistent across all states.

The partial region $ \mathcal{R} $ is defined as
\begin{equation} 
\mathcal{R} = \left\{ s \mid s = s_0 \text{ or } ((s' \to s) \in \mathcal{A}^*_v \text{ and } s' \in \mathcal{R}) \right\}
\end{equation} 
where the element is either the initial state or a state that can be transitioned to from another element via valid action. 
Obviously $ \mathcal{R} \subseteq \mathcal{S}$ for any sampling, and $\mathcal{R} = \mathcal{S}$ when the valid probability $p$ is set to $1$. 

In this way, $|\mathcal{R}|$ is much smaller, The expected ratio of the sizes of $|\mathcal{R}|$ and $|\mathcal{S}|$ satisfies the following formula,
\begin{equation} 
    \mathbb{E}[\frac{|\mathcal{R}|}{|\mathcal{S}|}]= \mathbb{E}[\frac{\sum_l |\mathcal{R}_l|}{\sum_l |\mathcal{S}_l|}] = \mathbb{E}[\frac{\sum_l p^l |\mathcal{S}_l|}{\sum_l |\mathcal{S}_l|}]
\end{equation} 

Here, $ l $ denotes the length of the trajectory from $ s_0 $ to $ s $. For simplicity, we refer to $ l $ as the length of $ s $, denoted by $ |s| $. The notation $ \mathcal{R}_l $ (or $ \mathcal{S}_l $) represents the set of all states with $ |s| = l $, where $ s \in \mathcal{R} $ (or $ s \in \mathcal{S} $). 
Through our method, $|\mathcal{R}|$ is effectively constrained. According to the above equation, the expected number of terminal states with trajectory length as $l$ is reduced to $p^l$. 
The small sizes of partial regions allow \modelname to efficiently assign appropriate flows to state transitions,  especially in scenarios with extremely large state spaces. 

In the following, we first detail how the planner decides whether to change  partial regions and how it selects partial regions. Then we introduce the partial local search algorithm which enhances the quality of generated candidates.


\subsection{Planner For \modelname}
The planner is responsible for selecting a partial region and determining when to transition from the current region to a new one. It comprises three key components: \textbf{selecting module}, \textbf{decision module} and \textbf{updating module}.

\subsubsection{Selecting Module}
This module handles the selection of partial regions. Various strategies can be employed for this purpose. Below, we introduce a heuristic strategy utilized in our implementation.

\textbf{Proportional Strategy}:
The Proportional Strategy selects a new region based on actions whose probabilities are proportional to their corresponding scores, which are given by $\mathrm{score}$ and updated through the updating module. Specifically, the probability of selecting the $i$-th action $a_i^*$ is given by,
\begin{equation}
    p(i) = \frac{\mathrm{score} (a_i^*)}{\sum_{j=1}^n \mathrm{score} (a_j^*)}
\end{equation}

Similar to Random Strategy, assume the current region is called $\mathcal{R}_0$ and the new region is called $\mathcal{R}_1$. Let $X_i$ be an indicator random variable representing the event where the i-th action $a_i^* \in \mathcal{A^*}$ is selected in both regions. However, considering the probability distribution $p$ is updated during sampling procedure, we use $p_1$ and $p_2$ to refer to the probability distribution of $\mathcal{R}_1$ and $\mathcal{R}_2$ and $p$ to refer to the probability of generating valid action.

The expected value of $X$ is:

\begin{equation}
\begin{aligned}
    \mathbb{E}[X] &= \mathbb{E}\left[\sum_{i=1}^n X_i\right] = \sum_{i=1}^n \mathbb{E}[X_i] = \sum_{i=1}^n p_1(i)p_2(i)n^2p^2
\end{aligned}
\end{equation}
The intersection of two regions is
\begin{equation}
\mathbb{E}[\mathcal{R}_1 \cap \mathcal{R}_2]  = \sum_l (\sum_{i=1}^n p_1(i)p_2(i))^ln^{2l}p^{2l}
\end{equation}
The union of two regions is
\begin{equation}
\mathbb{E}[\mathcal{R}_1 \cup \mathcal{R}_2] = \sum_l n^l(2p^l - \sum_l (\sum_{i=1}^n p_1(i)p_2(i))^ln^{l}p^{2l}) )
\end{equation}

The final indicator $I$ is
\begin{equation}
\begin{aligned}
I(p) = &\ \alpha_1 \frac{\sum_l \left( \sum_{i=1}^n p_1(i)p_2(i) \right)^l n^{2l} p^{2l}}{\sum_l n^l p^l} + \alpha_2 \frac{\sum_l n^l \left( 2p^l - \sum_l \left( \sum_{i=1}^n p_1(i)p_2(i) \right)^l n^l p^{2l} \right)}{\sum_l n^l p^l}
\end{aligned}
\end{equation}


\subsubsection{Updating Module}
The proportional strategy requires a score function to evaluate whether an action should be chosen. Here we introduce updating module which defines and updates the score function according to $bs$ and $rs$, where $ bs = (\tau_1, \tau_2, \dots, \tau_m) $ represents a batch of trajectories and $ rs = (r_1, r_2, \dots, r_m) $ represents the corresponding rewards of trajectories in the batch. Updating module updates the scores of actions in $ \mathcal{A}^* $ with the following strategy. Two auxiliary function $\mathrm{HR}$ $\mathrm{CNT}$ are needed to record historical rewards related to actions and the number of occurrences of actions. We traverse every action $a_i$ of every trajectory $\tau_j$ in $bs$, the updating strategy is formulated as,
\begin{equation}
    \begin{aligned}
        \mathrm{HR}(a_i) & = \mathrm{HR}(a_i) + r_j \\
        \mathrm{CNT}(a_i) & = \mathrm{CNT}(a_i) + 1 \\
        \mathrm{score}(a) &= \frac{\mathrm{HR}(a_i)+1}{\mathrm{CNT}(a_i)}
    \end{aligned} 
\end{equation}
Both $\mathrm{HR}(a_i)$ and $\mathrm{CNT}(a_i)$ are initialized with $0.01$ to prevent division by zero.

\subsubsection{Decision Module} 
The decision module evaluates the quality of recently sampled candidates relative to historical candidates. At the end of each training iteration, it decides whether to transition to a new partial region. The decision is made according to conditions related to whether the actor is exploring states in current iteration with higher scores (number of modes lastly found) than in historical iterations. If not so, it indicates that a transition to a new region is necessary, triggering the selecting module. The procedure is given as in Alg.ref\ref{alg:decision}. Note that we actually encourage the planner to transition to new regions, except in specific cases where the current region is clearly favorable.

\begin{algorithm}
\caption{Decision Change}
    {\bf Input:}
        $Step$: minimum step of explorations, $step$:the exploring step in current region, $iter$: the current exploring iteration, $His$: list of historical found number of modes, $\rm{avg}$: function computing the average exploring performance.
        $Diff$: difference array of $His$\\
     {\bf Output:} An indicator of whether to selecting a new region.
    \begin{algorithmic}[1]
    \IF {$step < Step$}
    \STATE {return False}
    \ENDIF 
    \IF {$Diff[iter] > \mathrm{avg}(His)\ \mathrm{or} \ Diff[iter] > Diff[iter-1]$}
    \STATE {return False}
    \ENDIF 
    \IF {$Diff[iter]+Diff[iter-1] > 2*\mathrm{avg}(His)$}
    \STATE {return False}
    \ENDIF
    \STATE {return True}
    \end{algorithmic}
    \label{alg:decision}
\end{algorithm}

\subsection{Partial Local Search}
The objective of actor is to explore as many high-scoring candidates as possible within a specific region before the planner chooses another region. Building upon Local Search GFlowNet~\cite{kim2023local}, which has been shown to consistently generate high-reward samples in scenarios with limited size of state spaces, we introduce two key adaptations to better integrate local search into our model:  
\textbf{i)} The forward policy $ P_F(\tau) $ is forced to sample exclusively from the valid action set $ \mathcal{A}_v = \mathcal{A}^*_v \cup \mathcal{A}' $. Such adaptation is required by the proposed Partial Search strategy.
\textbf{ii)} Rather than backtracking only $K$ steps, we backtrack the entire trajectory, which ensures the trajectories are valid.

Given a complete trajectory $(s_0, \cdots, s_{n-K}, \cdots, s_n)$ from initial state $s_0$ to terminal state $s_n$, we begin by backtracking $ K $ steps from the terminal state $s_n$ using $ P_B $ to obtain a new state $ s_{n-K}' $. Next, we reconstruct a terminal state $s'_n$ by applying a $ K $-step partial forward policy. Finally, we backtrack $ (n-K) $ steps from $ s_{n-K}' $ to generate a complete trajectory for $s'_n$, which can be denoted  as $(s_0, \cdots, s_{n-K}', \cdots, s'_n)$.

Following the construction in \cite{kim2023local}, our partial local search algorithm is an augmentation (\textbf{Step B} and \textbf{Step D}) of the standard algorithms for \gfn (\textbf{Step A} and \textbf{Step C}).

\subsubsection{\textbf{Step A}:Sampling}
\modelname samples a batch of complete trajectories $ (\tau_1, \tau_2, \dots, \tau_m) $, where each trajectory starts from $ s_0 $ and is generated using the forward policy $ P_F(\tau) $. The forward policy is constrained to explore only a specific region $ \mathcal{R} $, which is determined by the planner $ P $.

\begin{wrapfigure}{r}{0.5\textwidth}
   \includegraphics[width=0.48\textwidth]{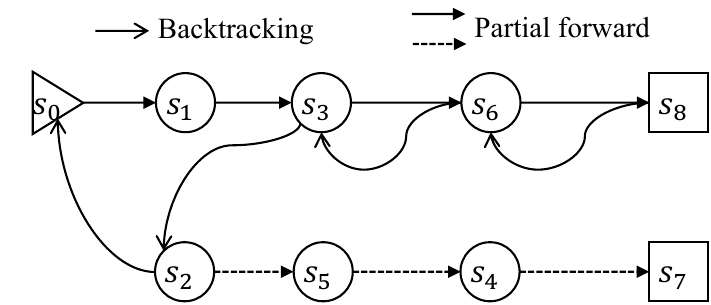}
   \caption{Illustration of a single round of trajectory reconstruction. The procedure repeats $I$ times for each originally sampled trajectory.}
   \label{fig:ls}
\end{wrapfigure}

\subsubsection{\textbf{Step B}:Refining}

In this step, the complete trajectories $ (\tau_1, \tau_2, \dots, \tau_m) $ sampled in \textbf{Step A} are reconstructed $ I $ times to explore high-scoring candidates. 

For each trajectory $ \tau = (s_0, s_1, s_2, \dots, s_n) $ in the complete trajectories $ (\tau_1, \tau_2, \dots, \tau_m) $, we perform a $ K $-step backtrack from the terminal state $ s_n $ using the backward policy $ P_B $, obtaining a partial trajectory $(s'_{n-K},\dots, s'_{n-1}, s_n)$. 
The $ K $-step backtrack process can be described as,
\begin{equation}
s_n \dashrightarrow s_{n-1}' \dashrightarrow \dots \dashrightarrow s_{n-K}'.
\end{equation}
Next, we reconstruct the trajectory by applying a $ K $-step forward policy starting from $ s_{n-K}' $, resulting in a new partial trajectory:
\begin{equation}
\tau_{\text{recon}} = (s'_{n-K}, \dots, s_n').
\end{equation}
Notice that the reconstructed trajectories are incomplete, as they starts from $ s'_{n-K} $ rather than $ s_0 $. The original partial trajectory $ (s_0, \dots, s_{n-K-1}) $ cannot be directly concatenated with $ (s'_{n-K}, \dots, s_n') $ unless $ s'_{n-K} = s_{n-K} $, as there is no transition from $ s_{n-K-1} $ to $ s'_{n-K} $. 

To achieve a complete and uniform reconstruction, we perform an additional $ (n-K) $-step backtrack starting from $ s'_{n-K} $, described as:
\begin{equation}
s'_{n-K} \dashrightarrow s'_{n-K-1} \dashrightarrow \dots \dashrightarrow s'_0.
\end{equation}
Since all trajectories share the same initial state, we have $ s'_0 = s_0 $. This yields the partial trajectory:
\begin{equation}
\tau_{\text{back}} = (s_0', \dots, s'_{n-K-1}).
\end{equation}
By concatenating the two partial trajectories, we obtain a new complete trajectory $ \tau' $:
\begin{equation}
\tau' = (\underbrace{s'_0, \dots, s'_{n-K-1}}_{\text{back}}, \underbrace{s'_{n-K}, \dots, s_n'}_{\text{recon}}).
\end{equation}
The above reconstruction procedure repeats $I$ times, and reconstructed trajectories with higher rewards than original trajectory are reserved for further training the actor.

\subsubsection{\textbf{Step C}:Training}
Several optimization objectives can be employed to train GFlowNets, including Flow Matching (FM)\cite{bengio2021flow}, Detailed Balance (DB)\cite{bengio2023gflownet}, Trajectory Balance (TB)\cite{malkin2022trajectory}, Sub-Trajectory Balance (subTB)\cite{pan2023better}. All these objectives can be utilized in the proposed \modelname.
As an example, the FM objective is defined as: 
\begin{equation}
   \mathcal{L}_{FM} (\tau) = \sum_{s \in \tau, s \neq s_0} \Bigg( \sum_{s: (s \rightarrow s') \in \mathcal{A}} F(s, s') - R(s') - \sum_{s'': (s' \rightarrow s'') \in \mathcal{A}} F(s', s'') \Bigg)^2
\end{equation}
The FM objective is computed over all trajectories $ \tau $, regardless of whether they are reconstructed. Notably, the objective is unrestricted, utilizing the full action set $ \mathcal{A} $ rather than the constrained set $ \mathcal{A}_v $.

\subsubsection{\textbf{Step D}: Updating}
The sampled trajectories are passed to the planner, which updates its strategy for selecting the next partial region. The planner evaluates the quality of the generated candidates and determines whether an immediate transition to a new region is warranted. If a transition is deemed necessary, the planner also identifies the specific region to which \gfn\ should transition.

\begin{figure}[t]
    \centering
    \includegraphics[width=\linewidth]{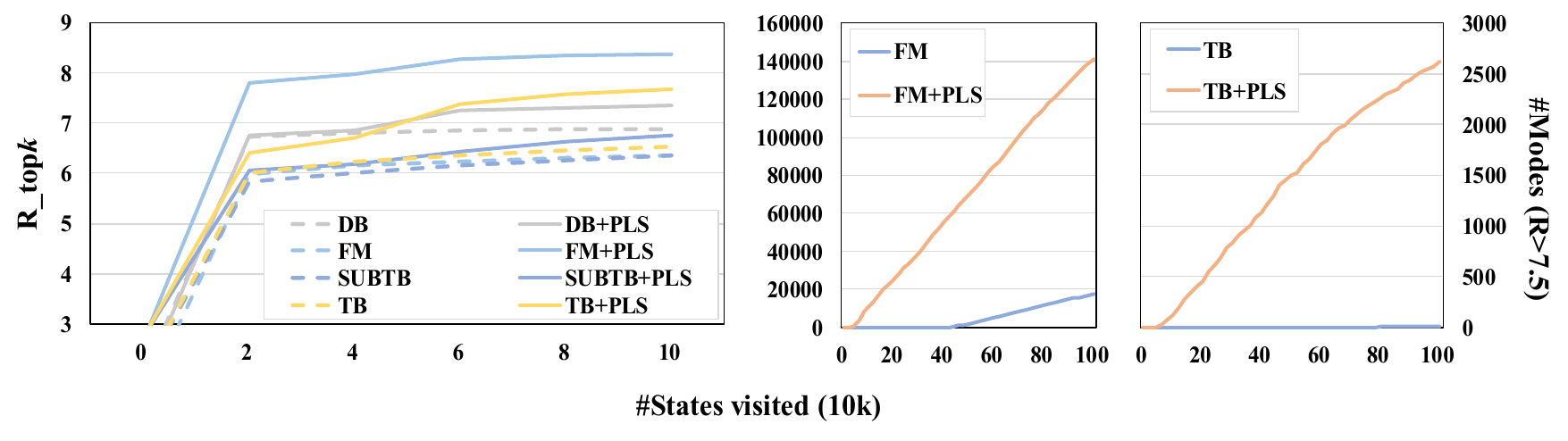}
    \caption{Training curves comparison with respect to R\_top$k$ and \#Modes (R$>$7.5) on the task of Molecule Design. }
    \label{fig:mols}
\end{figure}

\section{Experiment} \label{sec:exp}
In this section, we evaluate the proposed \modelname across three distinct tasks: Molecule Design, Sequence Generation, and RNA-Binding. Our evaluation primarily addresses two key research questions: 
1) How does the proposed framework facilitates the performance of existing GFlowNets with respect to both diversity and scoring of generated candidates? 
2) How do the different components of the framework affect performance? Especially partial search strategy.

To enable a comprehensive assessment, we select several task-specific metrics designed to evaluate both the \textit{scores} and \textit{diversity} of the generated candidates. All experiments are conducted on an NVIDIA Tesla A100 80GB GPU. To ensure a fair comparison, we retain the original experimental implementations and hyper-parameters without modification. Four \gfn\ objectives are implemented and evaluated across all experiments, including Flow Matching (FM), Detailed Balance (DB), Trajectory Balance (TB) and Sub-Trajectory Balance (SubTB). 

Four distinct configurations are evaluated: vanilla \gfn, \gfn\ with local search (LS), \gfn\ with partial search (P), and \gfn\ with partial local search (PLS). We provide detailed descriptions of the experimental settings and results for each task in the following sections.

\subsection{Molecule Design}

\subsubsection{Task definition}

In this task, the states are represented as molecule graphs or SMILES ~\footnote{\url{https://en.wikipedia.org/wiki/Simplified_Molecular_Input_Line_Entry_System}}, while the actions involve adding new components to the molecule. As such, the molecule design task can be framed as a decision process. We utilize a vocabulary of building blocks defined by junction tree modeling~\cite{jin2018junction}, which we inherit from the vanilla GFlowNet~\cite{bengio2021flow}. At each step, the action space is determined by two factors: selecting an atom to which a building block will be attached, and deciding which block to attach.

Notably, we increased the maximum allowed number of building blocks from $\textbf{8}$ in ~\cite{bengio2021flow} to $\textbf{12}$, which significantly expands the overall size of state space, with more than $\mathbf{10^{24}}$ terminal states. This enlargement leads to a much larger state space.


\begin{table}
  \centering
  \caption{Performance comparison on Molecule Design task.}
  \scalebox{0.8}{
    \begin{tabular}{ccc}
    \toprule
    Model & \#Modes (R$>$7.5) & R\_top$k$ \\
    \midrule
    \midrule
    DB    & 10    & 7.085 \\
    DB+LS    & 14    & 7.176 \\
    DB+P    & 56    & 7.521 \\
    DB+PLS & 78    & 7.606 \\
    \midrule
    FM    & 17898 & 8.319 \\
    FM+LS    & 21898    & 8.337 \\
    FM+P    & 124901    & 8.433 \\
    FM+PLS & \textbf{141115} & \textbf{8.508} \\
    \midrule
    subTB & 1     & 7.187 \\
    subTB+LS & 1  & 7.068 \\
    subTB+P & 11  & 7.351 \\
    subTB+PLS & 14    & 7.402 \\
    \midrule
    TB    & 8     & 7.290 \\
    TB+LS & 17  & 7.312 \\
    TB+P & 1333  & 7.951 \\
    TB+PLS & 2626  & 8.029 \\
    \midrule
    PPO   & 0     & 7.360 \\
    MARS  & 0     & 7.322 \\
    \bottomrule
    \end{tabular}%
  }
  \label{tab:mol}%
\end{table}
\subsubsection{Result}
\par To provide comprehensive benchmarking, we additionally include two baselines: MARS~\cite{xie2021mars} and PPO~\cite{schulman2017proximal}, enabling direct comparison between our \modelnames and established non-flow-based approaches. We introduce two key metrics. The first metric measures the number of diverse Bemis-Murcko scaffolds identified above a fixed reward threshold $T$. The second metric calculates the average reward of the top-$k$ distinct molecules discovered so far.

As reported in Tab.~\ref{tab:mol}, both MARS and PPO demonstrate limited exploration capabilities, failing to identify any high-scoring candidate throughout the optimization process. 
Among the GFlowNet objectives, TB, SubTB, and DB also fail, locating only a limited number of high-scoring modes in the final stages. 
Our proposed \modelname establishes new state-of-the-art performance, demonstrating a nearly tenfold improvement in mode discovery and candidate quality compared to vanilla FM-based GFlowNet approach. 
Furthermore, we show the training curves with respect to the two metrics in Fig.\ref{fig:mols}. \modelname demonstrates significantly enhanced exploration efficiency and convergence speed in vast state space, consistently identifying high-scoring candidates within remarkably few optimization steps. Overall, the large state space in this task causes a dramatic decline in the performance of the original GFlowNets. In contrast, the proposed \modelname effectively overcomes this challenge and converges towards high-scoring subregions in just a few steps.

\begin{figure}[h]
    \centering
    \includegraphics[width=\linewidth]{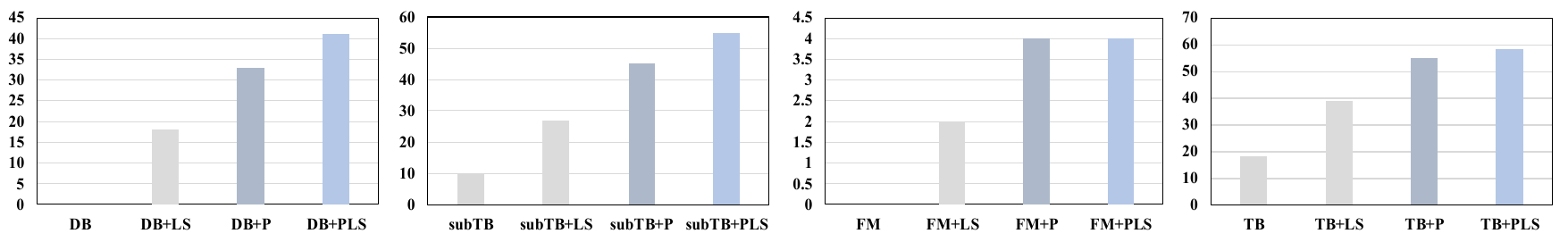}
    \caption{Performance comparison with respect to \#Modes on Sequence Generation task.}
    \label{fig:bitseq}
\end{figure}

\subsection{Sequence Generation}
\subsubsection{Task definition}
The task of sequence generation is to generate binary bit sequences using the set $\{0, 1\}$ with a fixed length $n=120$. The size of the state space is $2^{120} \approx 10^{36}$. The objective is to explore as many distinct modes as possible, where these modes are defined by a predefined list of sequences $M$. If the distance between a sampled candidate $s$ and a given predefined sequence $m \in M$ is smaller than a specified hyper-parameter threshold distance $d$, we consider the mode $m$ to have already been discovered. In this problem, we use Levenshtein Distance following \cite{malkin2022trajectory,zhang2023distributional}. The reward function for a sequence $s$ is defined as $R(s) = \max_{m \in M} \exp(-\mathrm{dist}(s,m))$. 

Since $M$ is a designated set of sequences, the upper bound on the number of discovered modes is restricted to $|M|$. $|M|$ is set to $60$ in our experiment.

Simply defining the fixed action set $\mathcal{A^*}$ as $\{0, 1\}$ is not feasible, because in that case, $\mathcal{A^*}$ would only have four subsets. Such a coarse granularity is unacceptable. A more reasonable approach is to fill $k$-bits at each step. With this approach, $\mathcal{A^*}$ operates at a finer granularity, having $2^{2^k}$ subsets depending on $k$. The bit sequence generation process is divided into $\frac{n}{k}$ positions. We define the action as $(a^*, a')$, where $a^*$ indicates which $k$-bits to choose, and $a'$ indicates which unfilled position to select.


\begin{wrapfigure}{r}{0.4\textwidth}
    \centering
    \includegraphics[width=\linewidth]{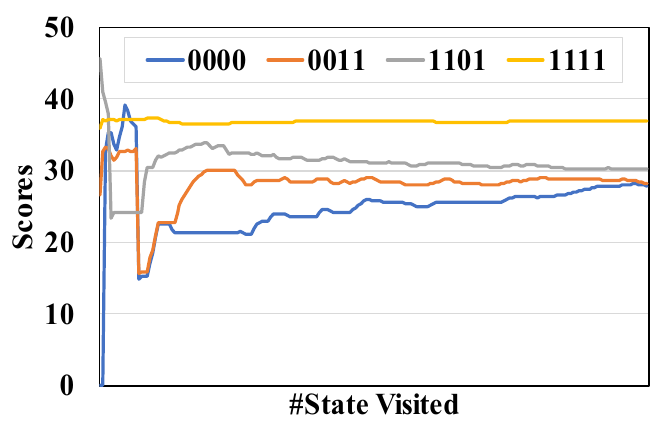}
    \vspace{-0.5cm}
    \caption{Performance comparison with respect to \#Modes on Sequence Generation task.}
    \label{fig:bitseq}
\end{wrapfigure}

\subsubsection{Result}
We report one metric of \modelnameNoBlank s, number of modes. Number of modes denotes how many modes the model finds, we define finding a mode by a edit distance less than 28. 

As shown in Fig.\ref{fig:bitseq}, through comprehensive comparative analysis, our \modelnames demonstrate significant improvements over prior GFlowNet implementations across all training objectives. Notably, \modelnames consistently outperform vanilla GFlowNets, demonstrating a great improvement in mode identification efficiency. Among the various training, the TB-based approach exhibits the most robust performance, establishing itself as the optimal configuration for our proposed framework in this task.


Additionally, to validate the effectiveness of the proposed strategy, we record the changing of scores of actions on bit sequence generation task to show what the scores look like and their convergence. The target sequences is generated by combining several basic sequences drawing from [10100101, 11111111, 11110000, 00001111, 00111100]. The actions are [0000\~1111]. The result shows that the scores of actions are converging during training. The score of '1111' is significantly higher than that of others, as it occurs more frequently than other actions in the basis bit sequences.

\begin{table*}[t]
  \centering
  \caption{Performance comparison on RNA-Binding task. The best results are bolded. \textbf{\textbackslash{}} means the model does not find any mode with reward greater than $0.95$. $k$ is set to $100$.}
  \scalebox{0.83}{
    \begin{tabular}{cc|cc|cc|cc|cc}
    \toprule
    Task  & \multicolumn{1}{c}{Metrics} & DB+PLS & \multicolumn{1}{c}{DB+LS} & TB+PLS & \multicolumn{1}{c}{TB+LS} & subTB+PLS & \multicolumn{1}{c}{subTB+LS} & FM+PLS & FM+LS \\
    \midrule
    \midrule
    \multirow{2}[2]{*}{RNA1} & \#Modes & 11    & 1     & \textbf{36} & 9     & 10    & \textbf{\textbackslash{}} & 11    & \textbf{\textbackslash{}} \\
          & R\_topk & 0.885  & 0.696  & \textbf{0.936 } & 0.875  & 0.886  & 0.668  & 0.858  & 0.695  \\
    \midrule
    \multirow{2}[2]{*}{RNA2} & \#Modes & 8     & \textbf{\textbackslash{}} & \textbf{18} & 8     & 10    & \textbf{\textbackslash{}} & 6     & \textbf{\textbackslash{}} \\
          & R\_topk & 0.878 & 0.660 & \textbf{0.923} & 0.893 & 0.879 & 0.686 & 0.853 & 0.682 \\
    \midrule
    \multirow{2}[2]{*}{RNA3} & \#Modes & 4     & 2     & \textbf{12} & 2     & 3     & \textbf{\textbackslash{}} & 2     & \textbf{\textbackslash{}} \\
          & R\_topk & 0.825 & 0.703 & 0.861 & \textbf{0.880} & 0.809 & 0.676 & 0.810 & 0.698 \\
    \bottomrule
    
      & \multicolumn{1}{c}{} & DB+P & \multicolumn{1}{c}{DB} & TB+P & \multicolumn{1}{c}{TB} & subTB+P & \multicolumn{1}{c}{subTB} & FM+P & FM \\
    \midrule
    \midrule
    \multirow{2}[2]{*}{RNA1} & \#Modes & 5    & \textbf{\textbackslash{}}     & 25 & 3   & 6    & \textbf{\textbackslash{}} & 4    & \textbf{\textbackslash{}} \\
          & R\_topk & 0.827  & 0.650  & 0.926 & 0.863  & 0.848  & 0.618  & 0.827  & 0.653  \\
    \midrule
    \multirow{2}[2]{*}{RNA2} & \#Modes & 4     & \textbf{\textbackslash{}} & 13 & 4   & 7    & \textbf{\textbackslash{}} & 2     & \textbf{\textbackslash{}} \\
          & R\_topk & 0.821 & 0.624 & 0.891 & 0.870 & 0.834 & 0.673 & 0.822 & 0.688 \\
    \midrule
    \multirow{2}[2]{*}{RNA3} & \#Modes & 2     & \textbf{\textbackslash{}}     & 7 & 1     & 2     & \textbf{\textbackslash{}} & 2     & \textbf{\textbackslash{}} \\
          & R\_topk & 0.814 & 0.696 & 0.845 & 0.817 & 0.812 & 0.667 & 0.802 & 0.669 \\
    \bottomrule
    
    \end{tabular}%
    }
  \label{tab:rna}%
\end{table*}%

\subsection{RNA-Binding}
\subsubsection{Task definition}
The goal of this task is to generate RNA sequences of 14 nucleobases. This process can be regarded as adding tokens to a string until it reaches the length of 14. There are 4 tokens: adenine (A),cytosine (C), guanine (G), uracil (U). We  use PA-MDP following the construction of \cite{kim2023local}. The reward $R$ is the binding affinity to the target transcription factor provided by \cite{lorenz2011viennarna}. We conduct three RNA binding tasks, L14-RNA1, L14-RNA2, and L14-RNA3 introduced by \cite{sinai2020adalead}.

In this task, the length of RNA sequence is fixed to 14 and the action size $|\mathcal{A}^*|$ is 4. Therefore, the state space is $4^{14}\approx 10^{8}$. This is definitely a small state space for GFlowNets. Even under this kind of circumstances, our proposed model shows faster convergence and better performance compared with existing methods.

\subsubsection{Result}
Following \cite{kim2023local}, we introduce two key metrics: the number of modes with reward greater than $0.95$ and top-$k$ performance. In the RNA-binding task, modes refer to high-scoring samples that surpass a predefined reward threshold and are distinctly separated according to a specified similarity constraint. Top-$ k $ performance measures the average reward of the top $ 100 $ highest-scoring sampled candidates. We run all the models for $2000$ rounds.

Through systematic comparison of methods with and without partial search capabilities, we demonstrate that partial search significantly enhances performance across multiple dimensions. Notably, it exhibits remarkable improvements in diversity metrics, even in limited state space scenarios. The method also achieves substantial gains in top-$k$ performance, establishing its comprehensive superiority over conventional approaches. 

\section{Conclusion} \label{sec:con}
Addressing the issue of significant performance degradation in large state space scenarios with existing GFlowNet methods, this paper proposes \modelnameNoBlank. \modelname introduces a novel planner which restricts the exploration in small partial state spaces. Due to the reduction in the size of the partial state spaces, the actor can more easily identify high-reward subregions. Several strategies are proposed to ensure the effectiveness of the selection of partial state spaces. Meanwhile, we modify local search algorithm to a partial version, enabling the deployment on partial state spaces, which further promotes the training of actor. The result of experiments on three tasks covering scenarios with large and small state spaces validates the effectiveness and efficiency of the proposed \modelnameNoBlank.

\bibliography{example_paper}
\bibliographystyle{plain}

\clearpage

\appendix

\section{Related Work} \label{sec:related}
\textbf{GFlowNet.} Since introduced by ~\cite{bengio2021flow}, GFlowNets have witnessed substantial research progress across both theoretical and applied domains. Theoretical advancements have primarily focused on three aspects: i) Connections to established methods, where ~\cite{malkin2022trajectory,zimmermann2022variational} established theoretical links between GFlowNets and variational methods, demonstrating superior performance over variational inference with off-policy training data; ii) Framework enhancements, including ~\cite{pan2022generative,pan2023better}'s work on improving credit assignment through intermediate signals, ~\cite{pan2023stochastic}'s integration of world modeling, and ~\cite{ma2024baking}'s solution to flow bias via isomorphism tests; iii) Theoretical foundations, with ~\cite{zhang2022unifying,lahlou2023theory,zhang2023diffusion} providing rigorous analyses and connecting GFlowNets to diffusion modeling.
In parallel, methodological innovations have emerged in several directions. ~\cite{zhang2022generative} pioneered the joint training of energy-based models with GFlowNets, introducing a bidirectional proposal mechanism that proved effective for discrete data modeling and was subsequently adapted by ~\cite{kim2023local} for local search algorithms. ~\cite{pan2023pre} contributed an unsupervised learning approach, while ~\cite{jain2022biological} explored multi-objective generation capabilities.
Practical applications have demonstrated GFlowNets' versatility, particularly in complex generation tasks. Notable implementations include biological sequence design by ~\cite{jain2022biological} and causal structure learning by ~\cite{deleu2022bayesian}. To address the computational challenges of candidate evaluation, current implementations commonly employ proxy models for efficient online assessment.
 
\textbf{Large Space Reinforcement Learning.} Reinforcement learning in large state or action spaces presents several significant challenges. Recent studies have proposed various approaches to address these issues. \cite{majeed2021exact} investigated methods for exact reduction of extensive action spaces through action sequentialization. \cite{zhang2021model} tackled the problem of maintenance optimization for degrading systems within large state spaces. \cite{ramesh2024distributionally} introduced a distributionally robust reinforcement learning method tailored for large state spaces. Additionally, \cite{yang2020function,wang2022tackling} explored the theoretical regret bounds of optimal algorithms, where regret is defined as the difference between the total rewards obtained by the agent and those achievable by an optimal agent. Despite these advancements, all existing methods are based on non-flow-based models. To the best of our knowledge, effective search strategies for flow-based models in large state spaces remain unexplored.

\section{More about Experiments}

\begin{wraptable}{r}{0.4\textwidth}
  \centering
  \caption{Training time per round.}
  \scalebox{0.8}{
    \begin{tabular}{ccccc}
    \toprule
    Method & FM    & DB    & SUBTB & TB \\
    \midrule
    Origin & 1.27  & 1.4   & 10.38 & 1.42 \\
    With PLS & 1.39  & 1.46  & 11.22 & 1.45 \\
    \midrule
    Delta & 0.094 & 0.043 & 0.081 & 0.021 \\
    \bottomrule
    \end{tabular}%
  }
  \label{tab:time}%
\end{wraptable}%

\subsection{Time consumption}
We introduced a planner to make high-level strategic decision-making, this inevitably brings some computation cost when the planner needs to update or decide. However, this cost is minor since it have nothing to do with the evaluation of the deep networks. In the table below, we report the average training time per round for the baseline models and their PLS variants, averaged for 100 rounds. As shown Tab.~\ref{tab:time}, The partial algorithm brings approximately $1/10$ more time when training with flow matching objective.

\subsection{RNA-Binding}
To provide a more comprehensive evaluation of \modelname's performance, we present additional results based on a broader set of metrics. These findings demonstrate that our \modelname outperforms prior baselines in various aspects.

\begin{figure}[h]
    \centering
    \includegraphics[width=0.32\textwidth]{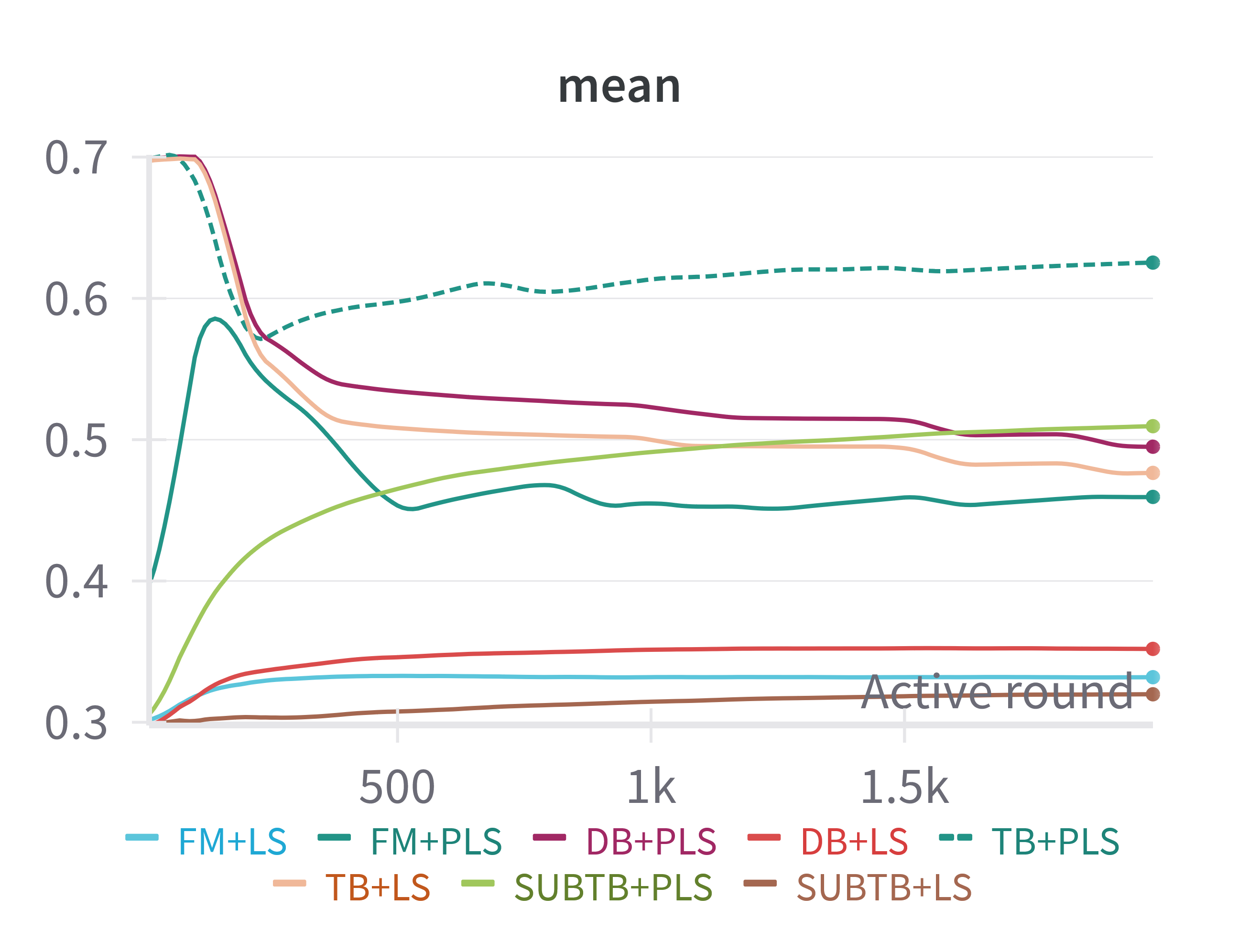}
    \includegraphics[width=0.32\textwidth]{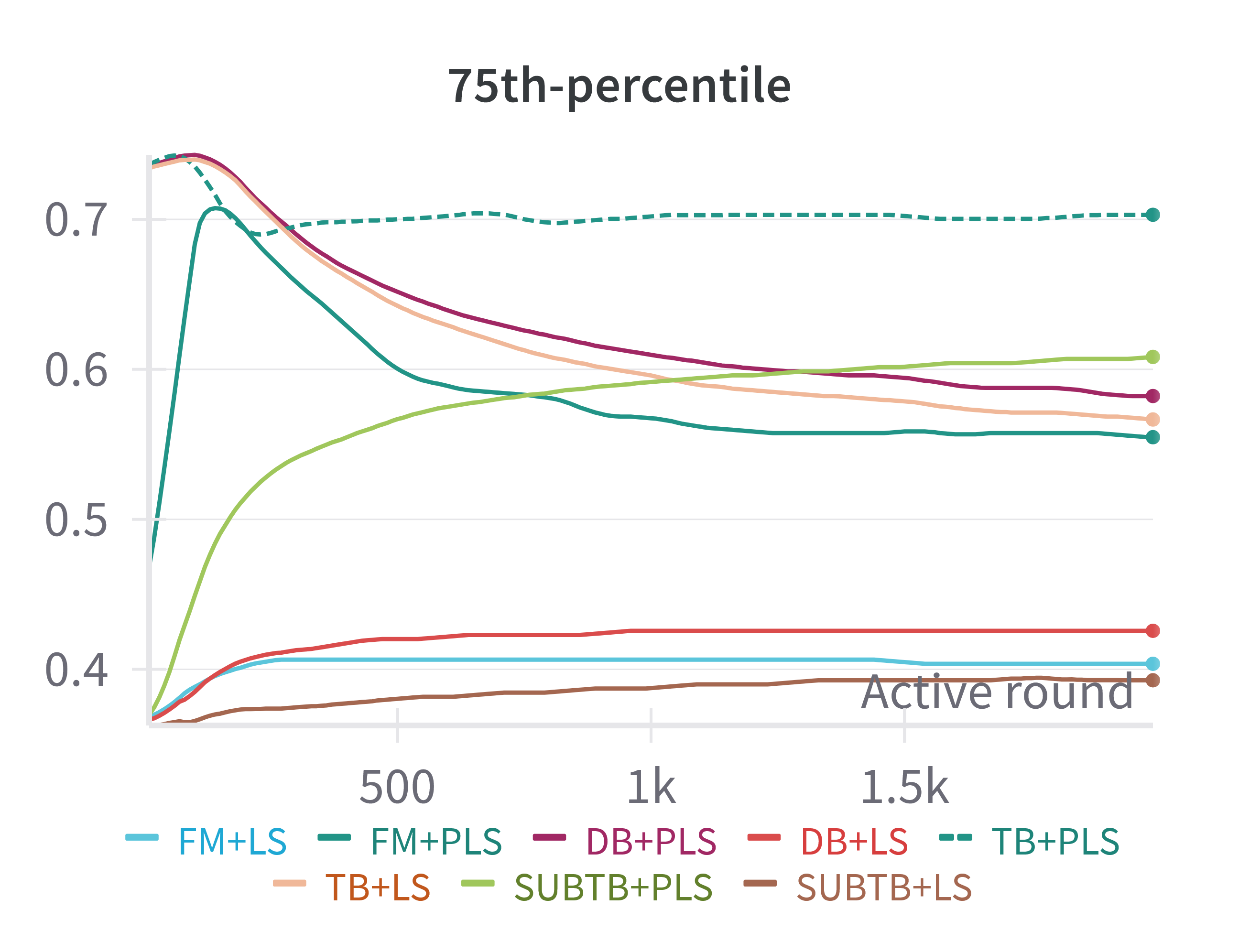}
    \includegraphics[width=0.32\textwidth]{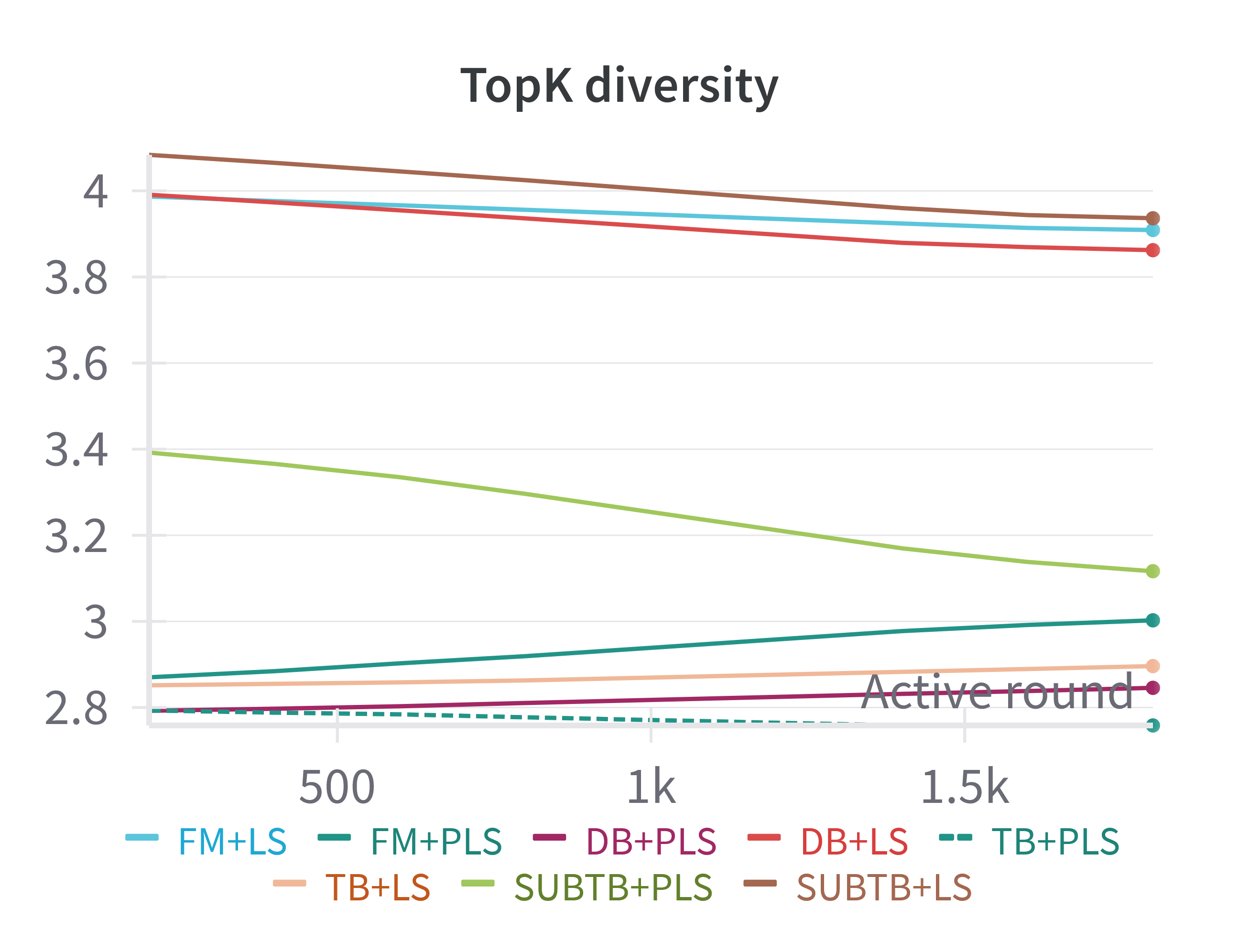}
    \caption{The RNA task 1}
\end{figure}

\begin{figure}[h]
    \centering
    \includegraphics[width=0.32\textwidth]{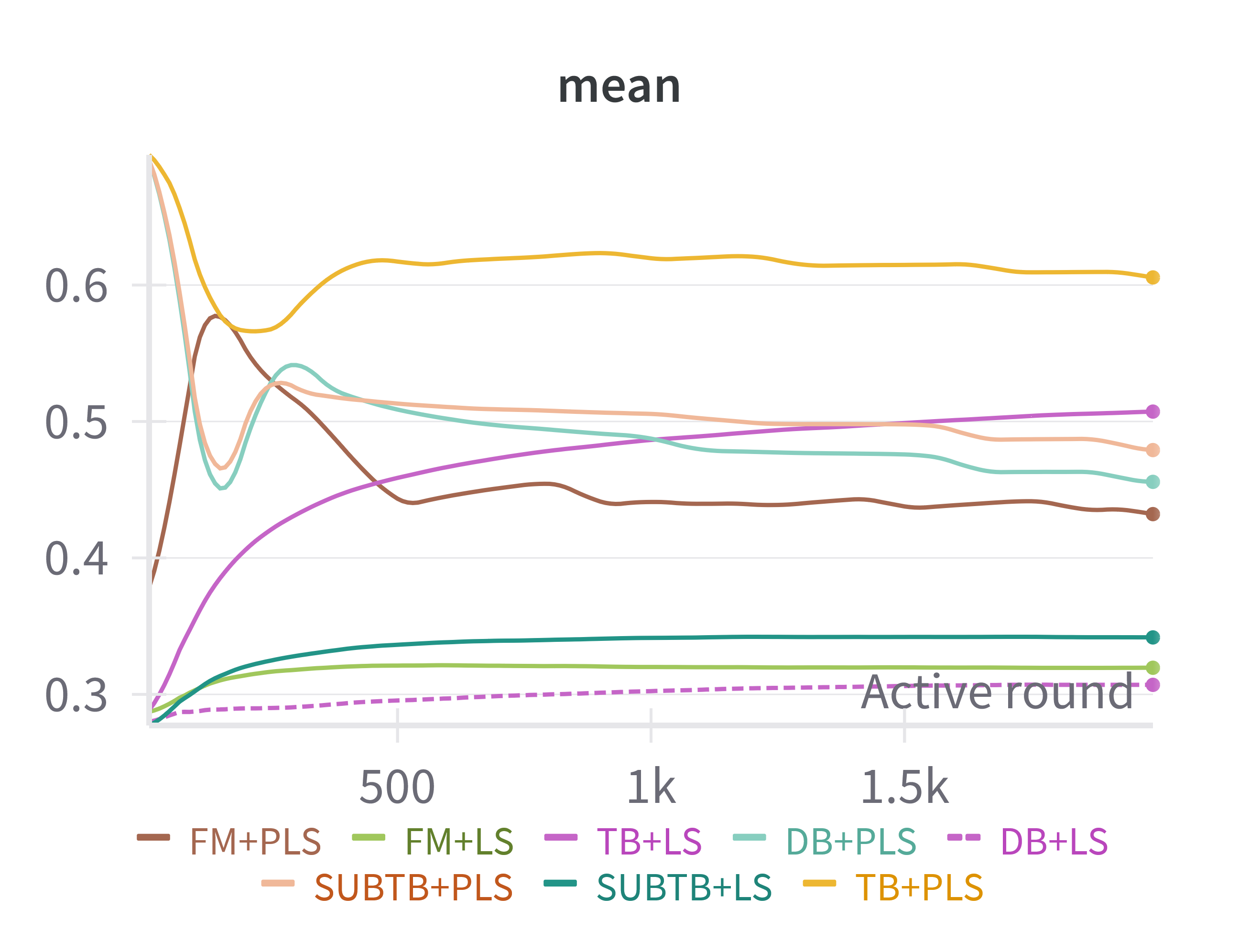}
    \includegraphics[width=0.32\textwidth]{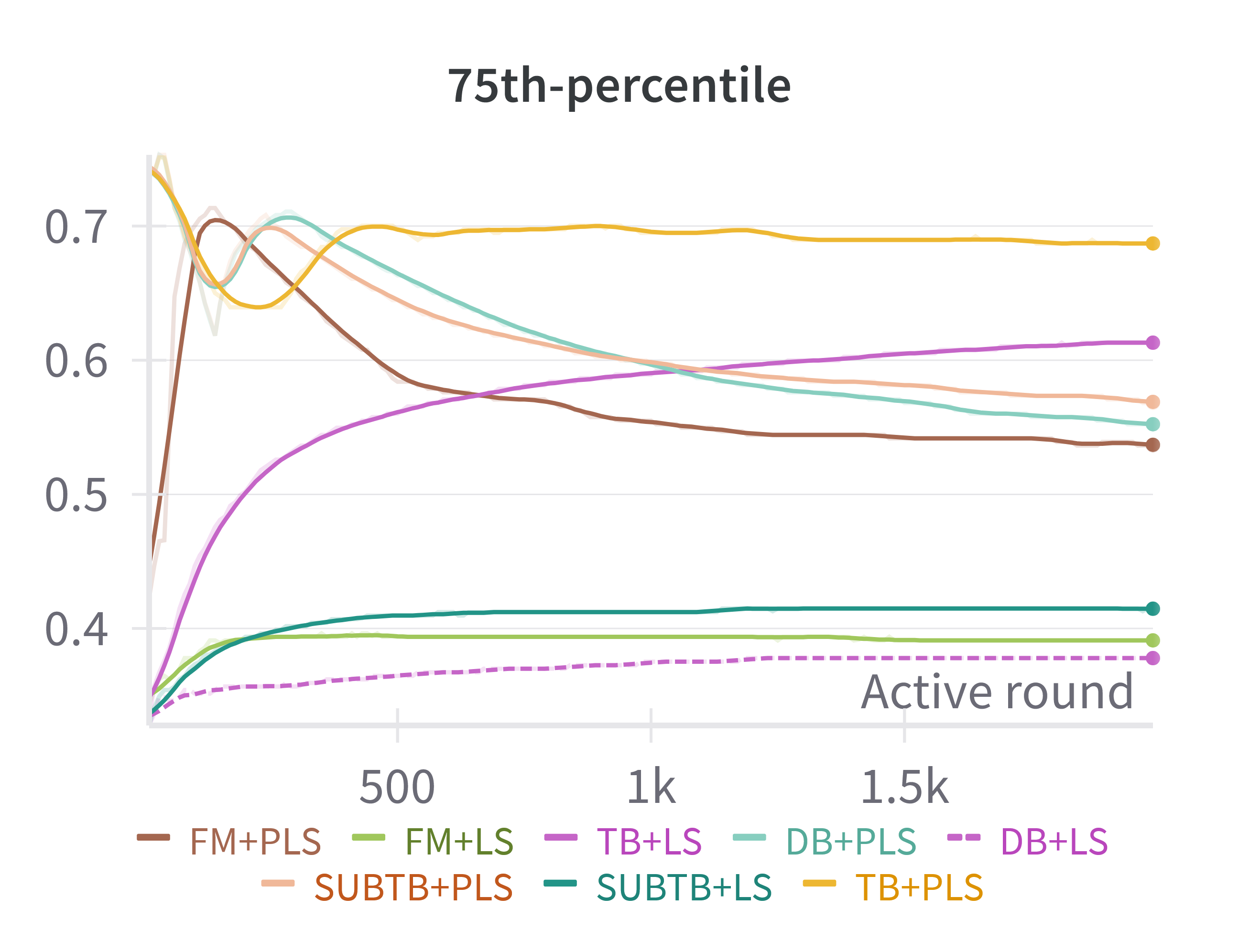}
    \includegraphics[width=0.32\textwidth]{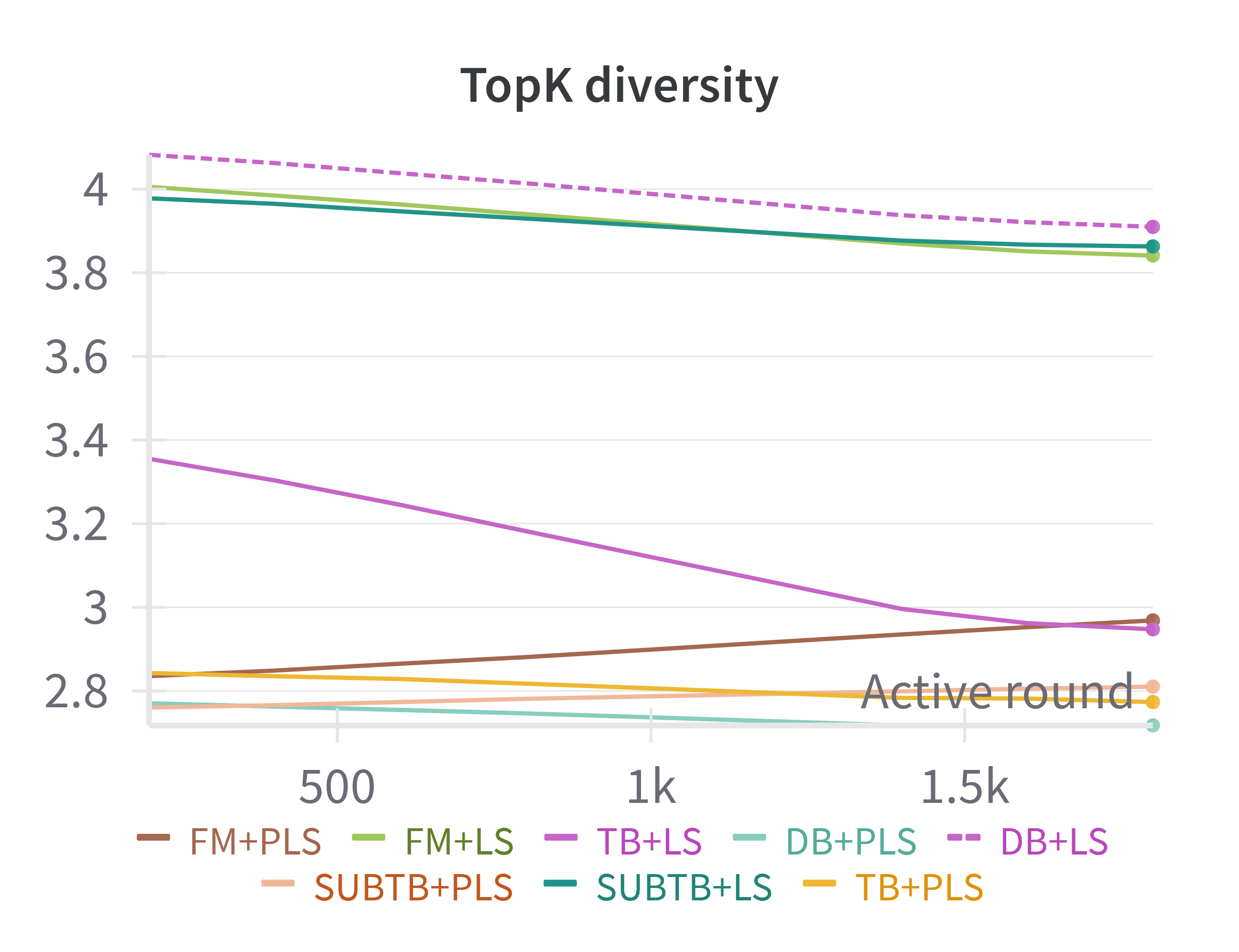}
    \caption{The RNA task 2}
\end{figure}

\begin{figure}[h]
    \centering
    \includegraphics[width=0.32\textwidth]{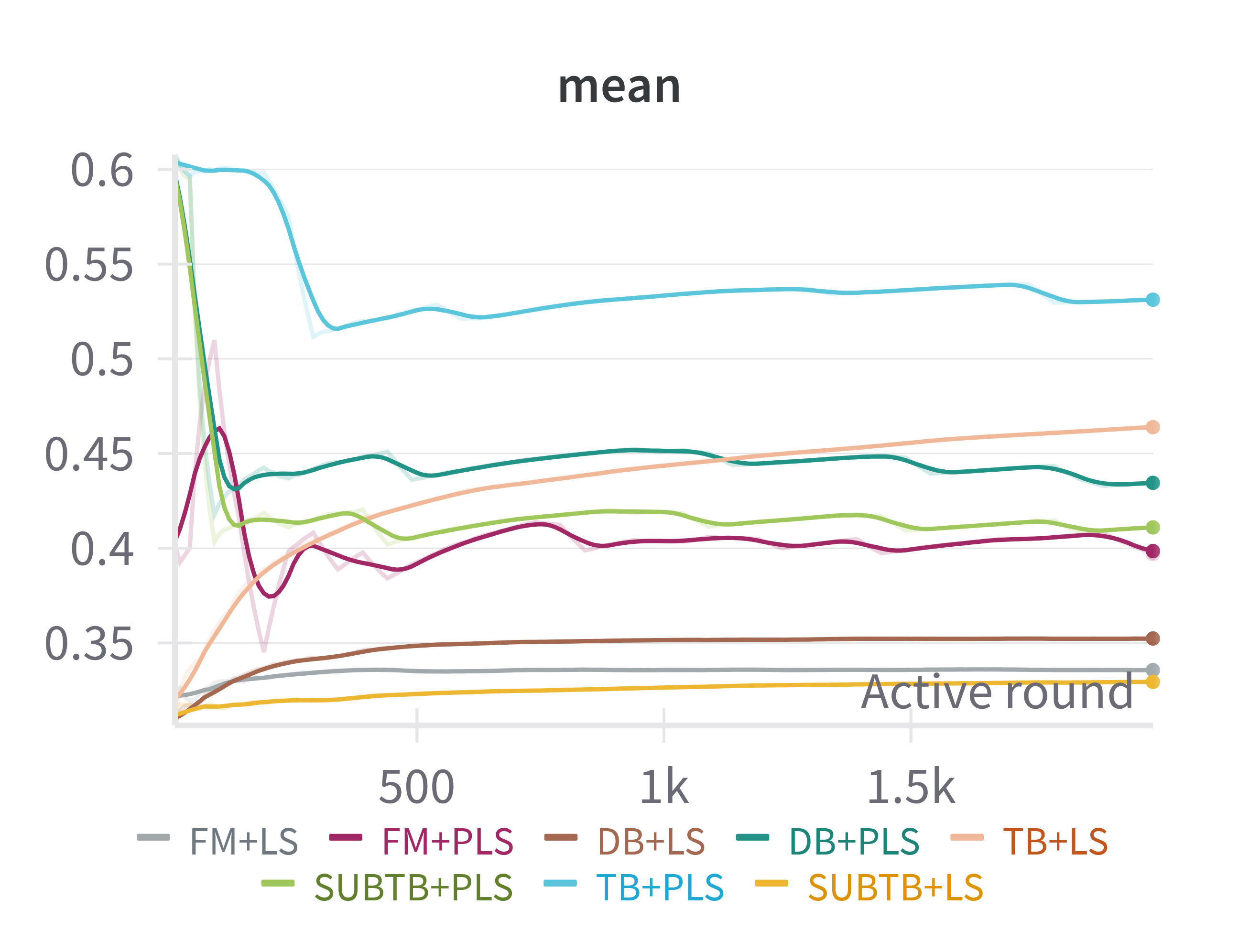}
    \includegraphics[width=0.32\textwidth]{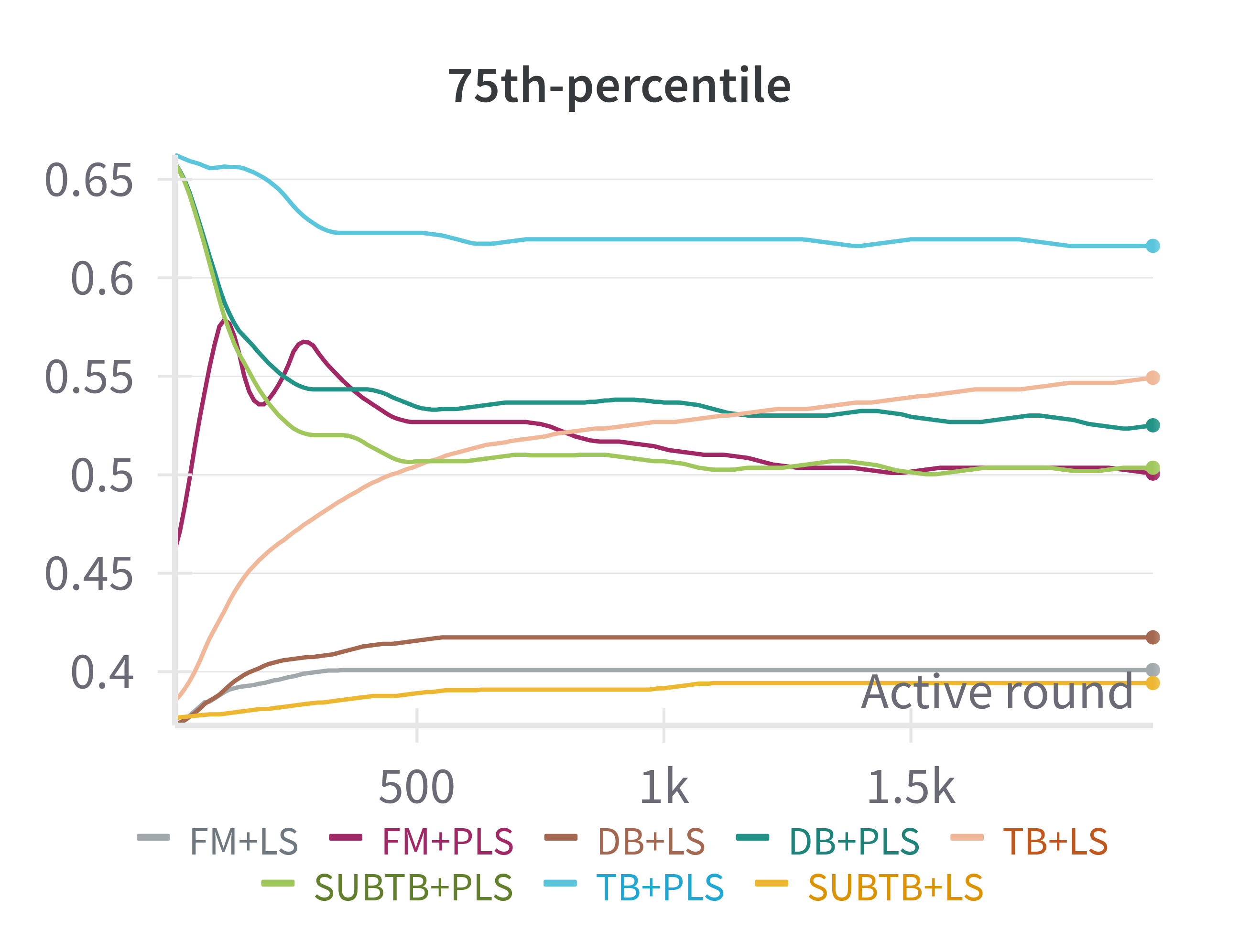}
    \includegraphics[width=0.32\textwidth]{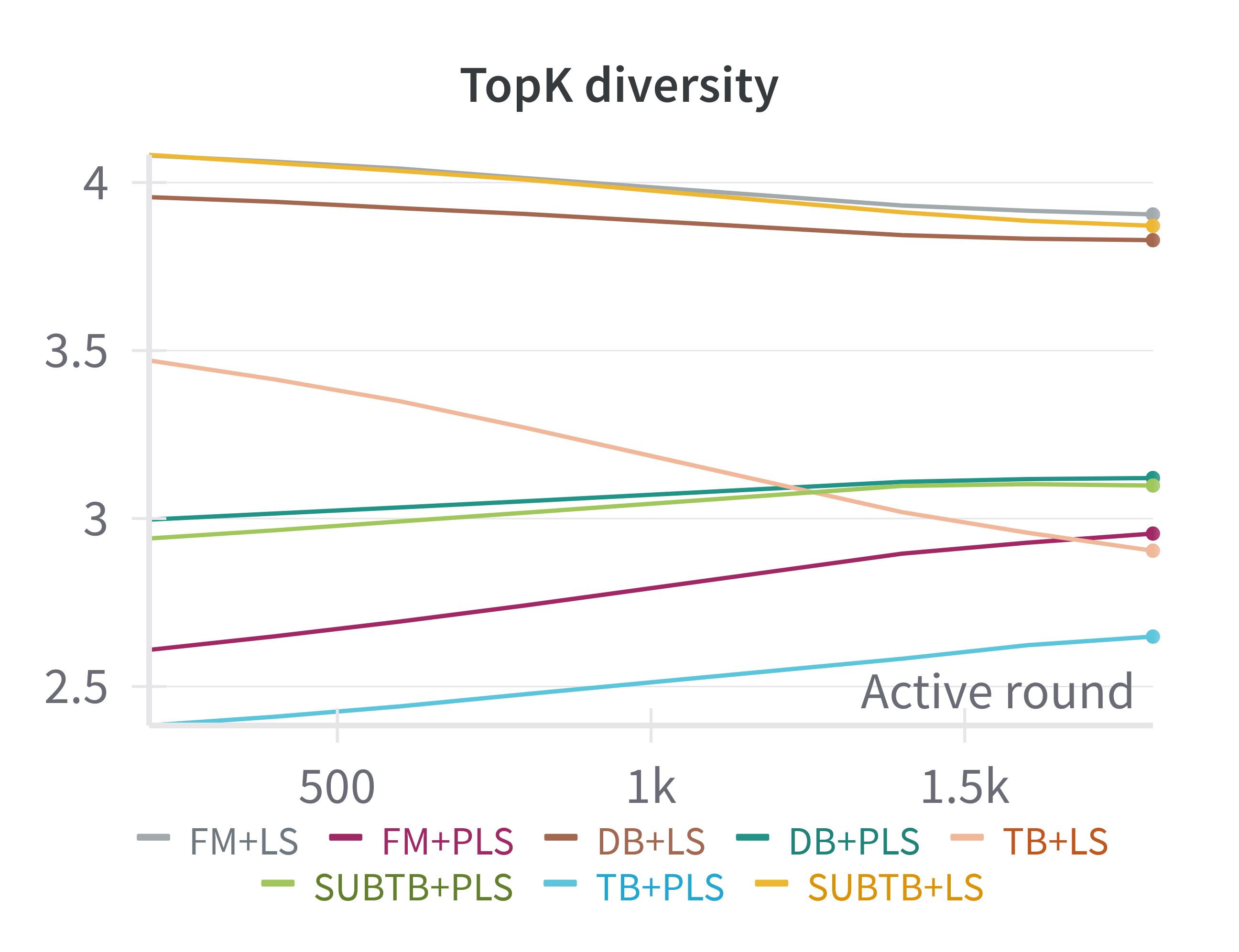}
    \caption{The RNA task 3}
\end{figure}

\section{Limitations}
Although our current methodology proves effective in many scenarios, it overlooks potential synergistic effects between blocks, particularly how specific combinations may consistently produce superior candidates due to their interactive properties. To overcome this limitation, we are investigating more advanced modeling approaches designed to capture these intricate inter-block relationships.

\newpage
\section*{NeurIPS Paper Checklist}

The checklist is designed to encourage best practices for responsible machine learning research, addressing issues of reproducibility, transparency, research ethics, and societal impact. Do not remove the checklist: {\bf The papers not including the checklist will be desk rejected.} The checklist should follow the references and follow the (optional) supplemental material.  The checklist does NOT count towards the page
limit. 

Please read the checklist guidelines carefully for information on how to answer these questions. For each question in the checklist:
\begin{itemize}
    \item You should answer \answerYes{}, \answerNo{}, or \answerNA{}.
    \item \answerNA{} means either that the question is Not Applicable for that particular paper or the relevant information is Not Available.
    \item Please provide a short (1–2 sentence) justification right after your answer (even for NA). 
\end{itemize}

{\bf The checklist answers are an integral part of your paper submission.} They are visible to the reviewers, area chairs, senior area chairs, and ethics reviewers. You will be asked to also include it (after eventual revisions) with the final version of your paper, and its final version will be published with the paper.

The reviewers of your paper will be asked to use the checklist as one of the factors in their evaluation. While "\answerYes{}" is generally preferable to "\answerNo{}", it is perfectly acceptable to answer "\answerNo{}" provided a proper justification is given (e.g., "error bars are not reported because it would be too computationally expensive" or "we were unable to find the license for the dataset we used"). In general, answering "\answerNo{}" or "\answerNA{}" is not grounds for rejection. While the questions are phrased in a binary way, we acknowledge that the true answer is often more nuanced, so please just use your best judgment and write a justification to elaborate. All supporting evidence can appear either in the main paper or the supplemental material, provided in appendix. If you answer \answerYes{} to a question, in the justification please point to the section(s) where related material for the question can be found.

IMPORTANT, please:
\begin{itemize}
    \item {\bf Delete this instruction block, but keep the section heading ``NeurIPS Paper Checklist"},
    \item  {\bf Keep the checklist subsection headings, questions/answers and guidelines below.}
    \item {\bf Do not modify the questions and only use the provided macros for your answers}.
\end{itemize}


\begin{enumerate}

\item {\bf Claims}
    \item[] Question: Do the main claims made in the abstract and introduction accurately reflect the paper's contributions and scope?
    \item[] Answer: \answerYes{} 
    \item[] Justification: \justificationTODO{}

\item {\bf Limitations}
    \item[] Question: Does the paper discuss the limitations of the work performed by the authors?
    \item[] Answer: \answerYes{} 
    \item[] Justification: \justificationTODO{}

\item {\bf Theory assumptions and proofs}
    \item[] Question: For each theoretical result, does the paper provide the full set of assumptions and a complete (and correct) proof?
    \item[] Answer: \answerYes{} 
    \item[] Justification: \justificationTODO{}

    \item {\bf Experimental result reproducibility}
    \item[] Question: Does the paper fully disclose all the information needed to reproduce the main experimental results of the paper to the extent that it affects the main claims and/or conclusions of the paper (regardless of whether the code and data are provided or not)?
    \item[] Answer: \answerYes{} 
    \item[] Justification: \justificationTODO{}

\item {\bf Open access to data and code}
    \item[] Question: Does the paper provide open access to the data and code, with sufficient instructions to faithfully reproduce the main experimental results, as described in supplemental material?
    \item[] Answer: \answerYes{} 
    \item[] Justification: \justificationTODO{}

\item {\bf Experimental setting/details}
    \item[] Question: Does the paper specify all the training and test details (e.g., data splits, hyperparameters, how they were chosen, type of optimizer, etc.) necessary to understand the results?
    \item[] Answer: \answerYes{} 
    \item[] Justification: \justificationTODO{}

\item {\bf Experiment statistical significance}
    \item[] Question: Does the paper report error bars suitably and correctly defined or other appropriate information about the statistical significance of the experiments?
    \item[] Answer: \answerNo{} 
    \item[] Justification: \justificationTODO{}

\item {\bf Experiments compute resources}
    \item[] Question: For each experiment, does the paper provide sufficient information on the computer resources (type of compute workers, memory, time of execution) needed to reproduce the experiments?
    \item[] Answer: \answerYes{} 
    \item[] Justification: \justificationTODO{}
    
\item {\bf Code of ethics}
    \item[] Question: Does the research conducted in the paper conform, in every respect, with the NeurIPS Code of Ethics \url{https://neurips.cc/public/EthicsGuidelines}?
    \item[] Answer: \answerYes{} 
    \item[] Justification: \justificationTODO{}

\item {\bf Broader impacts}
    \item[] Question: Does the paper discuss both potential positive societal impacts and negative societal impacts of the work performed?
    \item[] Answer: \answerNA{} 
    \item[] Justification: \justificationTODO{}
    
\item {\bf Safeguards}
    \item[] Question: Does the paper describe safeguards that have been put in place for responsible release of data or models that have a high risk for misuse (e.g., pretrained language models, image generators, or scraped datasets)?
    \item[] Answer: \answerNA{} 
    \item[] Justification: \justificationTODO{}

\item {\bf Licenses for existing assets}
    \item[] Question: Are the creators or original owners of assets (e.g., code, data, models), used in the paper, properly credited and are the license and terms of use explicitly mentioned and properly respected?
    \item[] Answer: \answerYes{} 
    \item[] Justification: \justificationTODO{}

\item {\bf New assets}
    \item[] Question: Are new assets introduced in the paper well documented and is the documentation provided alongside the assets?
    \item[] Answer: \answerNA{} 
    \item[] Justification: \justificationTODO{}

\item {\bf Crowdsourcing and research with human subjects}
    \item[] Question: For crowdsourcing experiments and research with human subjects, does the paper include the full text of instructions given to participants and screenshots, if applicable, as well as details about compensation (if any)? 
    \item[] Answer: \answerNA{} 
    \item[] Justification: \justificationTODO{}

\item {\bf Institutional review board (IRB) approvals or equivalent for research with human subjects}
    \item[] Question: Does the paper describe potential risks incurred by study participants, whether such risks were disclosed to the subjects, and whether Institutional Review Board (IRB) approvals (or an equivalent approval/review based on the requirements of your country or institution) were obtained?
    \item[] Answer: \answerNA{} 
    \item[] Justification: \justificationTODO{}

\item {\bf Declaration of LLM usage}
    \item[] Question: Does the paper describe the usage of LLMs if it is an important, original, or non-standard component of the core methods in this research? Note that if the LLM is used only for writing, editing, or formatting purposes and does not impact the core methodology, scientific rigorousness, or originality of the research, declaration is not required.
    \item[] Answer: \answerNA{} 
    \item[] Justification: \justificationTODO{}

\end{enumerate}

\end{document}